\documentclass[conference]{IEEEtran}

\usepackage{lipsum}
\usepackage{graphicx}
\usepackage{subfigure}
\usepackage{amsmath}

\makeatletter
\def\ps@headings{%
	\let\@oddhead\@empty
	\let\@evenhead\@empty
	\def\@oddfoot{\@IEEEheaderstyle\hfil\thepage}%
	\def\@evenfoot{\@IEEEheaderstyle\thepage\hfil\hbox{}}%
}
\def\ps@IEEEtitlepagestyle{%
	\def\@oddhead{\strut\hfill2019 5\textsuperscript{th} IEEE International WIE Conference on Electrical and Computer Engineering (WIECON-ECE)\strut\hfill}
	\def\@evenhead{\strut\hfill2019 5\textsuperscript{th} IEEE International WIE Conference on Electrical and Computer Engineering (WIECON-ECE)\strut\hfill}
	\def\@oddfoot{\footnotesize 978-1-7281-4499-3/19/\$31.00~\copyright2019 IEEE\hfill\thepage}%
	\let\@evenfoot\@empty
}

\makeatother
\begin{document}
	
	\title{Breast Cancer Histopathology Image Classification and Localization using Multiple Instance Learning}
	
\author{\IEEEauthorblockN{Abhijeet Patil}
\IEEEauthorblockA{\textit{Department of Electrical Engineering} \\
\textit{Indian Institute of Technology Bombay}\\
Mumbai, India \\
abhijeetptl@ee.iitb.ac.in}
\and
\IEEEauthorblockN{Dipesh Tamboli}
\IEEEauthorblockA{\textit{Department of Electrical Engineering} \\
\textit{Indian Institute of Technology Bombay}\\
Mumbai, India \\
dipesh@ee.iitb.ac.in}
\and
\IEEEauthorblockN{Swati Meena}
\IEEEauthorblockA{\textit{Department of Electrical Engineering} \\
\textit{Indian Institute of Technology Bombay}\\
Mumbai, India \\
swatimeenasm@ee.iitb.ac.in}
\and
\hspace{1.0in}
\IEEEauthorblockN{Deepak Anand}
\IEEEauthorblockA{\textit{\hspace{1.0in}Department of Electrical Engineering} \\
\textit{\hspace{1.0in}Indian Institute of Technology Bombay}\\
\hspace{1.0in}Mumbai, India \\
\hspace{1.0in}
deepakanand@ee.iitb.ac.in}
\and
\IEEEauthorblockN{Amit Sethi}
\IEEEauthorblockA{\textit{Department of Electrical Engineering} \\
\textit{Indian Institute of Technology Bombay}\\
Mumbai, India \\
asethi@ee.iitb.ac.in}
}
	
	\maketitle 

	\begin{abstract}
		Breast cancer has the highest mortality among cancers in women. Computer-aided pathology to analyze microscopic histopathology images for diagnosis with an increasing number of breast cancer patients can bring the cost and delays of diagnosis down. Deep learning in histopathology has attracted attention over the last decade of achieving state-of-the-art performance in classification and localization tasks. The convolutional neural network, a deep learning framework, provides remarkable results in tissue images analysis, but lacks in providing interpretation and reasoning behind the decisions. We aim to provide a better interpretation of classification results by providing localization on microscopic histopathology images. We frame the image classification problem as weakly supervised multiple instance learning problem where an image is collection of patches i.e. instances. Attention-based multiple instance learning (A-MIL) learns  attention on the patches from the image to localize the malignant and normal regions in an image and use them to classify the image. We present classification and localization results on two publicly available BreakHIS and BACH dataset. The classification and visualization results are compared with other recent techniques. The proposed method achieves better localization results without compromising classification accuracy.
	\end{abstract}


\section{Introduction}
Breast cancer is the most prominent cause of death in women, and the number of breast cancer cases are increasing throughout the world \cite{spanhol_breast_2016}. Expert pathologists are required to perform a diagnosis of breast cancer, which is time-consuming. Pathologists make their decision based on various visual features observed in pathology slides such as morphological features of nuclei, micro, and macrostructure of nuclei, etc. Computer-aided diagnosis (CAD) systems can help pathologists to make decisions automatically. These techniques can also reduce inter-observer variations to make the diagnosis process reproducible.

Deep learning algorithms have produced performance at par with human experts on image classification and object detection tasks \cite{yamashita_convolutional_2018}. The convolutional neural network is the most widely used deep learning framework to learn complex discriminative features between image classes. Various architectures of CNNs such as VGG16 \cite{zhang_accelerating_2015} and ResNet18 \cite{he_deep_2015} have produced exceptional results on the massive ImageNet dataset over the years. CNNs are being used on medical images to produce state-of-the-art results.

Deep neural networks are often criticized for their lack of interpretability. In most of the application, neural networks act as black box feature extraction units. Lack of interpretability is more serious in applications like medical image analysis. Visualization of CNN features is an active area of research and techniques like guided backpropagation \cite{springenberg_striving_2014}, deconvolution \cite{springenberg_striving_2014}, and CAM-related methods propose localization for the predicted class in an input image. Grad-CAM \cite{selvaraju_grad-cam:_2016} gives localization for all the classes present in image. All of these techniques track gradients flowing in the backward pass of CNNs to produce localization in an input image, which results in localization over most prominent features representing a class.


Gradient-based localization techniques do not produce good results on histopathology images as features are distributed over most of the part of the image. In this study, we use attention-based multiple instance learning \cite{ilse_attention-based_2018} to produce better localization of malignant regions in breast histopathology images. We frame the image classification problem as a weakly supervised learning problem by assigning a single label to several patches of an input image. More specifically, we make a bag from multiple instances (patches) of an image and use attention-based multiple instance learning (A-MIL) for classification. 

In A-MIL, instance pooling assigns a learned weight to each instance to aggregate features to a bag~\cite{ilse_attention-based_2018} These learned weights can be used for localization as each weight signifies the importance of a particular patch for the classification task. We overlay these weights on an input image to show our localization results. A-MIL produces similar classification results compared to widely used CNN architectures such as ResNet18 and VGG16. A-MIL is shown to perform better in localization task as compared to Grad-CAM without compromising the accuracy of the classification task.

\section{Related work}
Pathologists grade cancer by observing the micro and macro-structures present in the histopathology slides. This task is very repetitive but critical for the treatment and diagnosis. CAD has come a long way to aid pathologist in these decisions. Conventional approaches in CAD involves feature-extraction based on the texture and appearances of the nuclei and its micro-environment~\cite{filipczuk_computer-aided_2013,george_remote_2014,zhang_breast_2011}. These features include perimeter, compactness, smoothness, eccentricity, solidity, equivalent diameter, extent, major axis length, and minor axis length of the nuclei and the texture features of surrounding areas. These features are fed to analysis methods like fuzzy-C-means, Gaussian mixture models, SVM, MLP and clustering algorithms to decide the class for the histopathology entities like nuclei or patch. These methods were popular with small datasets, however with large dataset they fail to generalize.

Deep learning models are well suited to the cases with large amount of data. These methods can learn intricate features from the histopathology slides and generalize well across patients, disease conditions, hospitals and are even robust to human-induced errors in the slide preparation. Deep learning based approaches are often based on CNNs~\cite{spanhol_breast_2016,golatkar_bach}.

Though the classification of histopathology images using deep learning algorithms has been explored extensively, very little work has been done on visualization or localization on histopathology datasets. Gradient-based methods such as guided-backpropagation \cite{springenberg_striving_2014}, deconvolution \cite{springenberg_striving_2014}, Grad-CAM \cite{selvaraju_grad-cam:_2016} provide good visualization on natural images but fail to provide reasonable localization on histopathology images due to challenges of large size, variation across disease states, and human-induced errors in slide preparation. We used attention-based multiple instance learning approach for classification and localization of relevant areas in histopathology images.

\section{Dataset and Methodology}
\subsection{Datasets}
We used publically available datasets, BreakHis~\cite{spanhol_breast_2016}, and BACH \cite{noauthor_iciar2018-challenge_nodate} for our analysis. BreakHis dataset contains 7909 images of four different magnification levels divided into two major classes \textit{viz.} benign and malignant. Each image in BreakHis dataset is of size $700 \times 460$. Magnification level-wise samples in each class is shown in Table \ref{tab:breakhis_dataset}. We split datasets into 80\% for training and 20\% for validation at each magnification to perform the experiments.

\begin{table}
\caption{Summary of BreakHis dataset}
\label{tab:breakhis_dataset}
\begin{center}
\begin{tabular}{|c|c|c|c|}
    \hline
    Magnification factor & Benign & Malignant & Total\\ 
    \hline
    40$\times$ & 652 & 1,370 & 1,995\\
    \hline
    100$\times$ &	644  &	1,437  &	2,081\\
    \hline
200$\times$  &	623  &	1,390 & 2,013\\
\hline
400$\times$  &	588  &	1,232  &	1,820\\
\hline
Total of Images &	2,480 &	5,429 &	7,909 \\
\hline
\end{tabular}
\end{center}
\end{table}

\begin{figure}[ht] 
\centering
  \subfigure[Benign]{%
    \includegraphics[width=4cm]{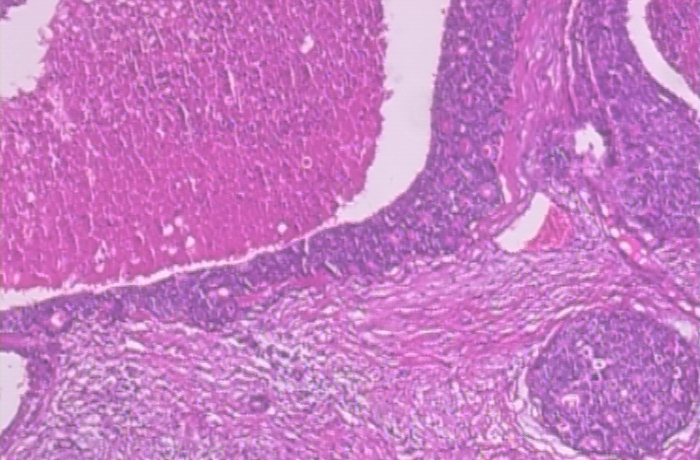} 
  } 
  \quad
  \subfigure[Malignant]{%
    \includegraphics[width=4cm]{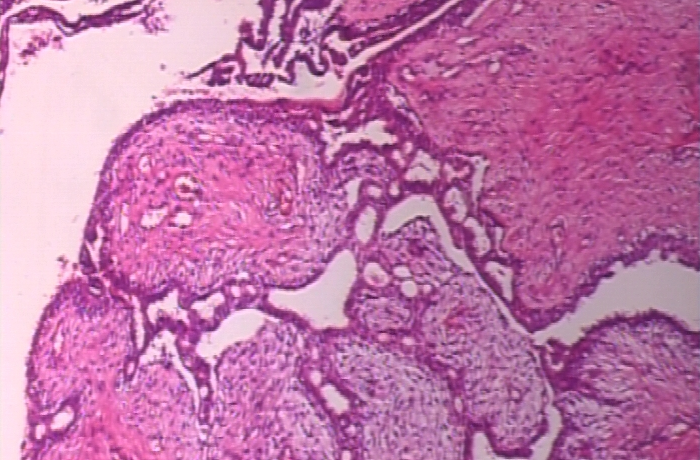}
  } 
  \subfigure[Benign]{%
    \includegraphics[width=4cm]{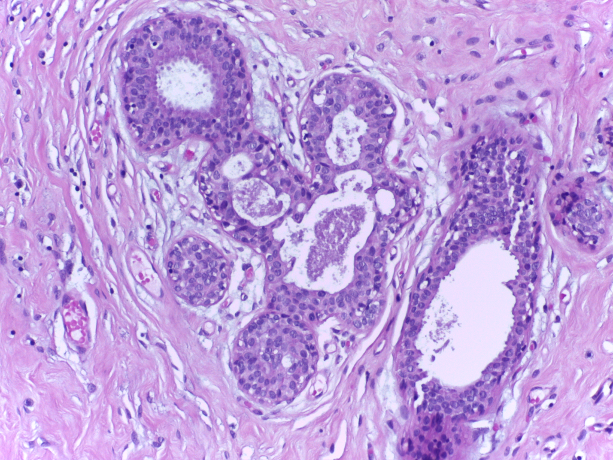} \label{Benign} 
  } 
  \quad
  \subfigure[Invasive]{
    \includegraphics[width=4cm]{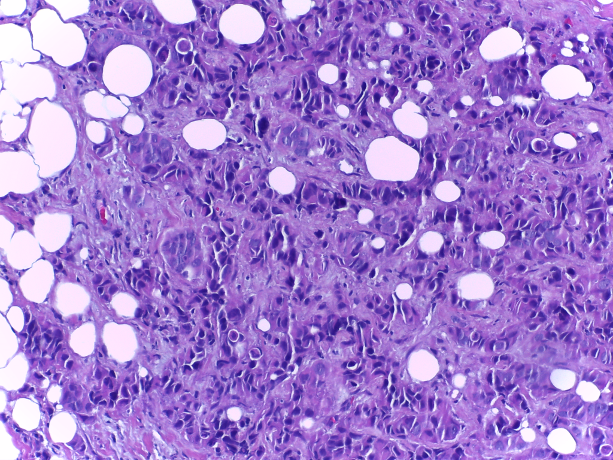} \label{Malignant} 
  } 
  \caption{Representative images from BreakHis and BACH datasets.} 
  \label{fig:breakhis_images}
\end{figure}   

The second dataset was ICIAR2018 Grand Challenge on Breast Cancer Histology images (BACH). The BACH dataset comprises of 400 histopathology images of breast cancer. Each image of this dataset is of three channels and the size of 2048 $\times$ 1536 pixels. The original BACH dataset contains four classes $viz.$ normal, benign, in situ and invasive. We clubbed normal and benign classes to form one class for our binary classification problem, whereas the other class is formed by clubbing in situ and invasive classes together. Figure \ref{fig:breakhis_images} shows representative images from BreakHis and BACH dataset.

\subsection{Multiple Instance Learning (MIL)}
Multiple Instance Learning (MIL) provides the solution for a weakly supervised learning problem. In MIL, the task is to predict a classification label of a bag, which consists of multiple instances. If $\mathcal{D}$ = \{$X_1$, $X_2$, ..., $X_n$\}, where $X_i$'s are the bags in dataset $\mathcal{D}$ and one bag contains $m$ instances i.e. $X_1$ = \{$i_1$, $i_2$, ..., $i_m$\} where $i_j$ is $j$th instance with a binary label $y_{i}$ in a bag $X_i$. We say that bag $X_i$ is positive if at least one instance $i_j$ is positive in bag $X_i$ as shown in equation \ref{eq:1}.

\begin{equation}
  Y = \left \{
  \label{eq:1}
  \begin{aligned}
    &0, && \text{if}\ \Sigma_m y_m=0 \\
    &1, && \text{otherwise}
  \end{aligned} \right.
\end{equation} 

The most critical part of MIL is the instance-level pooling. Instance level pooling aggregates instance level features to obtain bag level features. The most popular instance pooling operations in MIL are the mean pool and max pool. Mean pool operations averages over all the instances to predict the bag label, whereas max pool operations takes the maximally activated instance label as the bag label. Both max pooling and mean pooling have their disadvantages. Max pooling only accounts for the maximum activation which may be an outcome of an outlier. On the other hand, mean pooling weighs ever instance equally thus losing the information from the sparsely populated classes. In the method that we selected, instance level features $h_1$,$h_2 $, $h_3$, .., $h_m$ are pooled by taking their weighted average as shown in equation \ref{eq:2}~\cite{ilse_attention-based_2018}. The coefficients of weighted average pooling are learned using a two-layer neural network with softmax activation. Equation \ref{eq:3} gives the expression for attention computation.

\begin{equation}
    \label{eq:2}
    z={\sum_{p=1}^{m}} a_p h_p
\end{equation}
where,
\begin{equation}
   \label{eq:3}
    a_p = \frac{\exp{\{w^T \text{tanh}(Vh_p^T)\}}}{\Sigma_{j=1}^m \exp{\{w^T \text{tanh}(Vh_j^T)\}}}
\end{equation}
and $w \in \mathcal{R}^{l \times 1}$ and $V \in \mathcal{R}^{l \times m}$. In the above equation $l$ is number of instance level features and $a_p$ is the attention weight learned by the network.

\subsection{Bag Preparation}
We divided each three channel $700 \times 460$ image in the BreakHis dataset in $28 \times 28$ patches with stride 28. This resulted in 400 patches per image, which we used as instances in a single bag. In the BACH dataset, we took patches of $124 \times 124$ from the original image of $2408 \times 1536$ that gives 192 patches per image.

\subsection{A-MIL framework}
\begin{figure}[ht]
\includegraphics[scale=0.25]{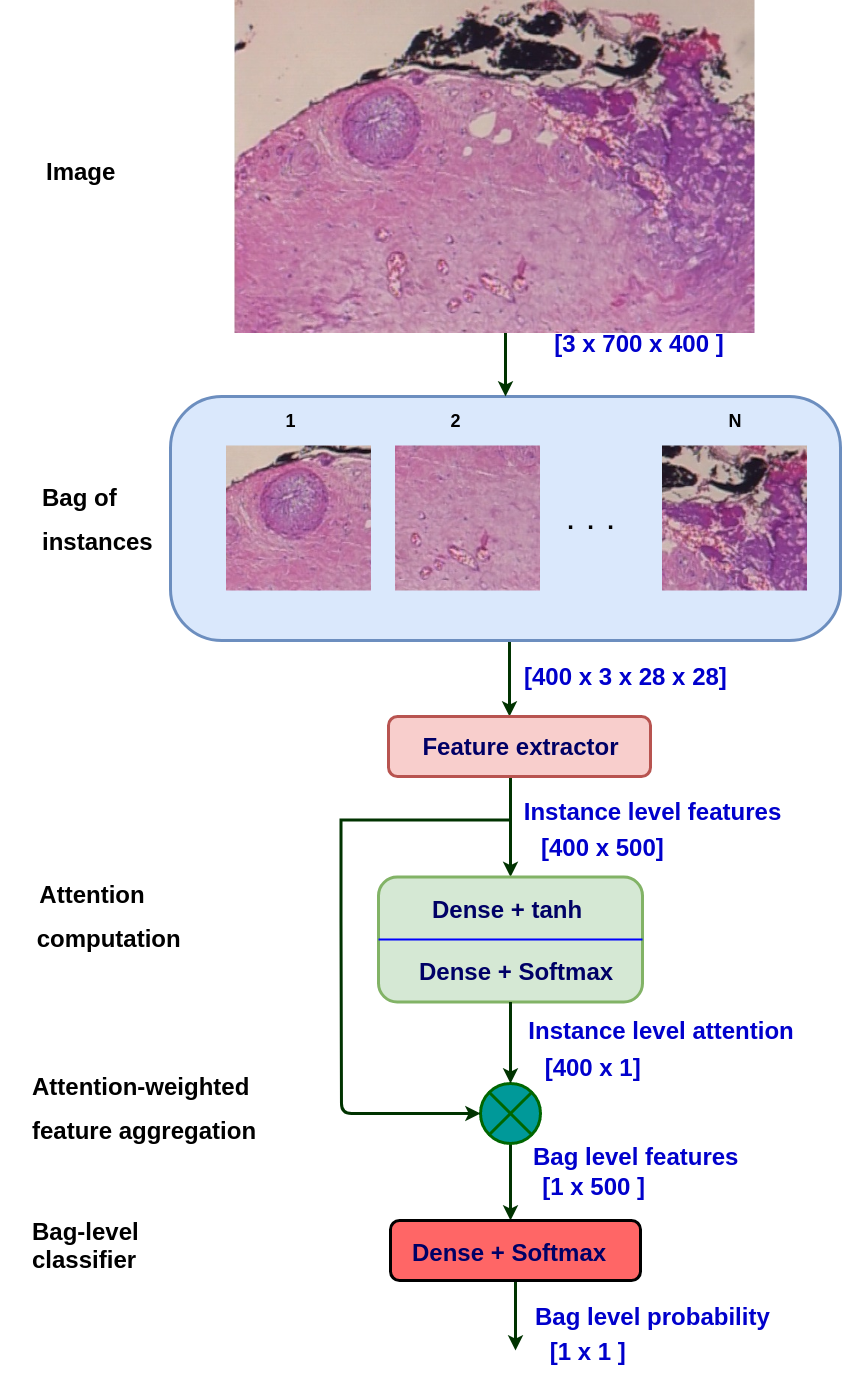}
\caption{Attention MIL architecture}
\label{fig:AMIL}
\end{figure}

The overall pipeline of A-MIL framework is shown in Figure \ref{fig:AMIL}, which is inspired by~\cite{ilse_attention-based_2018}. Each patch in a bag is processed through a feature extractor to get instance level features. The dense layer extracts 500 features from each instance. The attention computation block computes attention score using these 500 features of each instance. These attention weights are further used for attention aggregation to get the bag level features. A-MIL allows different weights for different instances in a bag. The attention aggregation computation makes the bag highly informative for the bag-level classifier. The detailed architecture of feature extractor used in A-MIL is given in Table \ref{feature_table}.

We trained the A-MIL framework with a batch size of one, the learning rate of $0.001$ and binary cross-entropy as a loss function. We used data augmentation such as vertical and horizontal flip, rotation by $90^{\circ}$, $180^{\circ}$, and $270^{\circ}$.

\begin{table}[ht]
\caption{Feature extractor used in A-MIL}
\label{feature_table}
\begin{center}
\begin{tabular}{ |c|c|} 
 \hline
 Input dimension & Layer\\ 
 \hline
 [$3 \times 28 \times 28$] & Conv1\\
 \hline
 [$20 \times 24 \times 24$] & Maxpool\\
 \hline
 [$20 \times 12 \times 12$] & Conv2\\
 \hline
 [$50 \times 8 \times 8$] & Maxpool\\
 \hline
 [$50 \times 4 \times 4$] & Flatten\\
 \hline
 800 & FC\\
 \hline
 500 & Extracted features\\
 \hline
 \end{tabular}
\end{center}
\end{table}     

\section{Experiments and Results}
We compared our classification results with commonly used transfer learning techniques. We trained a VGG16 network initialized with ImageNet trained weights. For training of VGG16, we took random patches of size $224 \times 224$ for training. We used the same strategy to train ResNet18 model. Heavy data augmentation is used for training these networks as the number of tunable parameters is large in VGG16 and ResNet18. We also trained a custom neural network with a lesser number of parameters. This custom convolutional neural network contains five convolutional layers, followed by two fully connected (FC) layers. The detailed architecture of the custom network is given in Table \ref{tab:CustomNet}.

Comparative classification results for four models is shown in Table \ref{tab:accuracy}. A-MIL technique gives similar performance compared to transfer learning on VGG16 and ResNet18. A-MIL also gives similar results to customNet, which has a comparable number of parameters to A-MIL network.

\begin{table}[ht]
\caption{Details of CustomNet architecture}
\label{tab:CustomNet}
\begin{center}
\begin{tabular}{ |c|c|} 
 \hline
 Input dimension & Layer\\ 
 \hline
 [$3 \times 128 \times 128$] & Conv1\\
 \hline
 [$64 \times 128 \times 128$] & Maxpool\\
 \hline
 [$64 \times 64 \times 64$] & Conv2\\
 \hline
 [$64 \times 64 \times 64$] & Maxpool\\
 \hline
 [$64 \times 32 \times 32$] & Conv3\\
 \hline
 [$128 \times 32 \times 32$] & Maxpool\\
 \hline
 [$128 \times 16 \times 16$] & Conv4\\
 \hline
 [$128 \times 16 \times 16$] & Maxpool\\
 \hline
 [$3 \times 8 \times 8$] & Conv5\\
 \hline
 [$128 \times 4 \times 4$] & Flatten\\
 \hline
 2048 & FC1\\
 \hline
 1024 & FC2\\
 \hline
\end{tabular}
\end{center}
\end{table}

\begin{table}[ht]
\caption{Classification accuracy by different models on BreakHis dataset}
\label{tab:accuracy}
\begin{center}
\begin{tabular}{ |c|c|c|c|c| } 
 \hline
 Networks /
Accuracy & $40\times$ & $100\times$ & $200\times$ & $400\times$\\ 
 \hline
 customNet & 80.56 & 82.13 &  79.11 &83.89\\
 \hline
 VGG16\_pretrained & 78.64 & 78.88 & 78.57  &  72.16\\ 
 \hline
 ResNet18\_pretrained &  83.6 & 82.58 & 85.11  & 84.23 \\ 
 \hline
 A-MIL & 82.95 & \textbf{86.45} &   \textbf{86.56}  & \textbf{84.43} \\ 
 \hline
\end{tabular}
\end{center}
\end{table}

\begin{figure}
 \centering 
 \includegraphics[width=8cm]{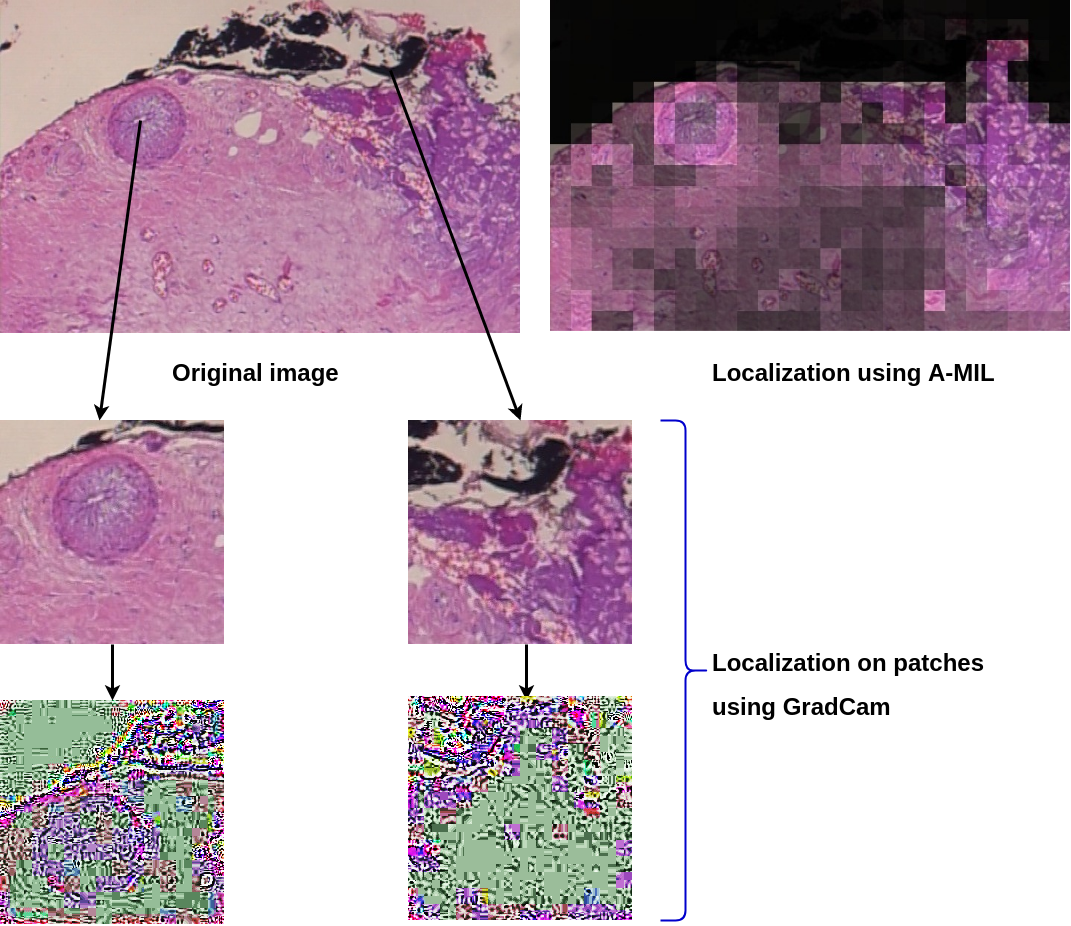}
 \caption{Comparison of visualization by A-MIL and Grad-CAM} 
 \label{fig:breakhis_localization}
\end{figure}

For localization, we overlay the attention weights on the input image and compared these localization results with Grad-CAM. We used Grad-CAM on VGG16 model trained on BreakHis dataset. Figure \ref{fig:breakhis_localization} demonstrates comparison between the localization results using A-MIL and Grad-CAM. We have shown Grad-CAM visualization on a patch containing tumor gland and on a background patch. Grad-CAM can only detect edges in input images, whereas A-MIL produces better visualization. A-MIL accurately localizes the malignant gland and ignores the background, as shown in Figure \ref{fig:breakhis_localization}. 

\begin{figure}[ht] 
\centering
  \subfigure[Original image]{%
    \includegraphics[width=4cm]{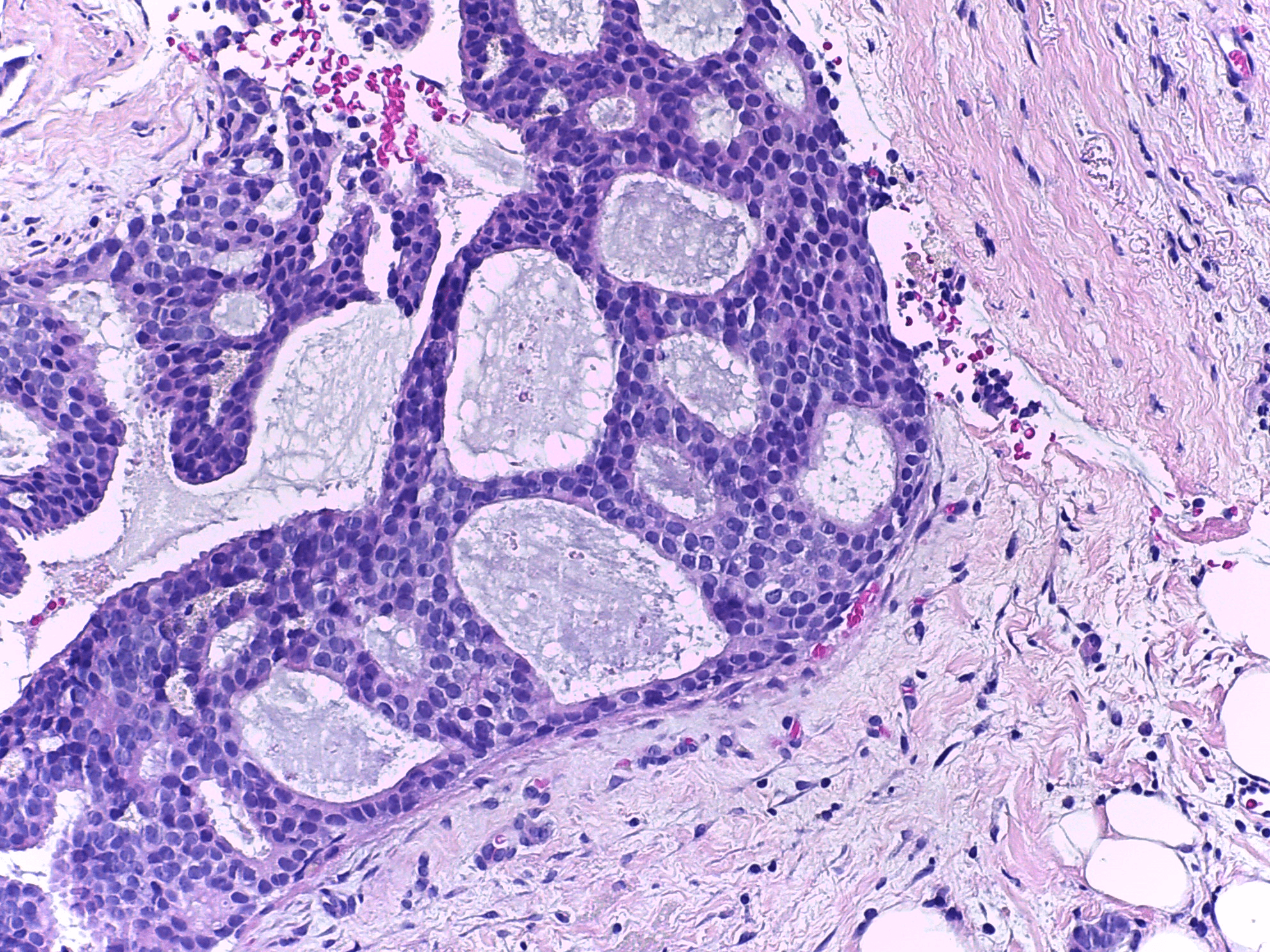}
  } 
  \quad
  \subfigure[Localization by A-MIL]{
    \includegraphics[width=4cm]{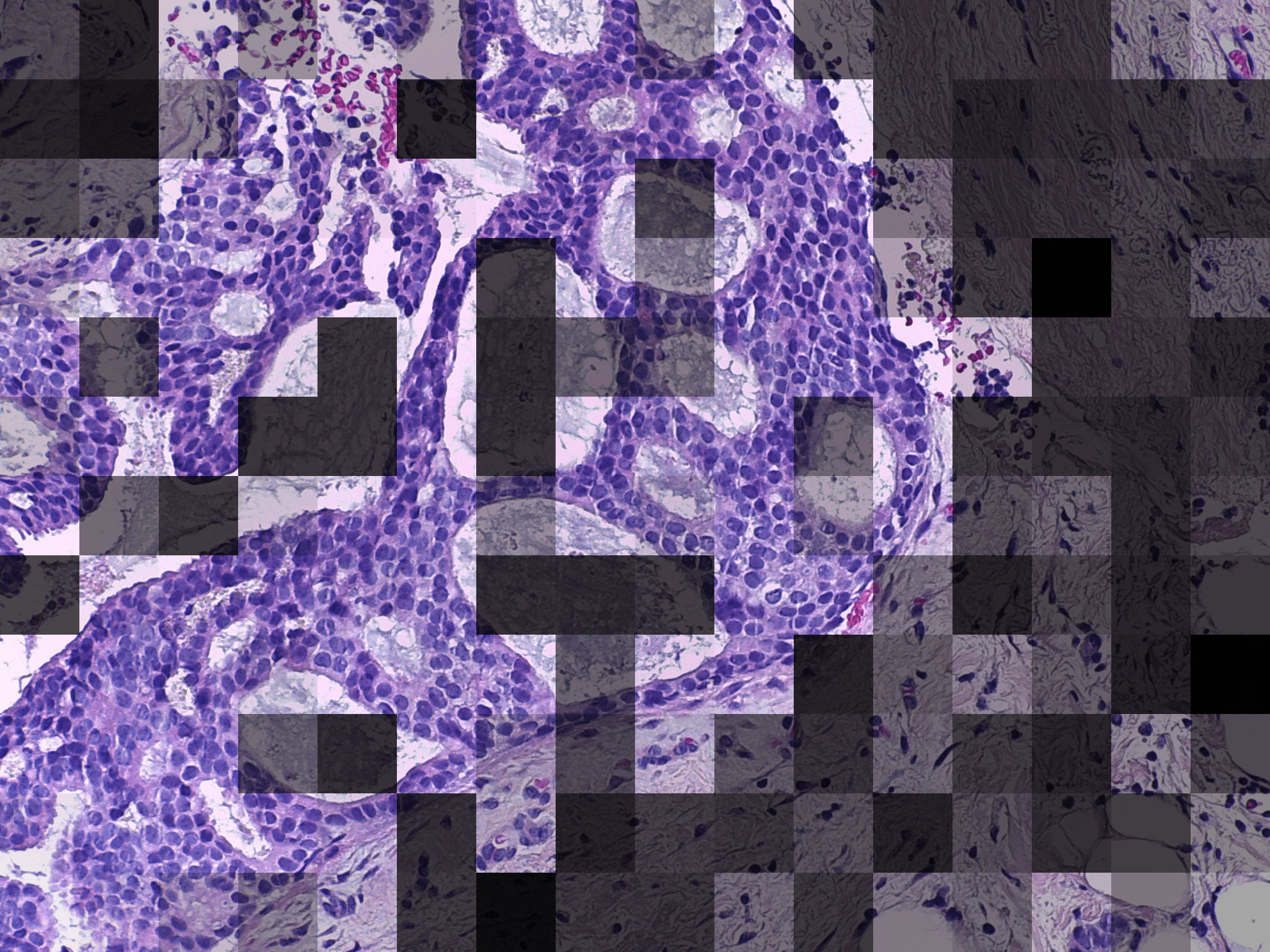} 
  } 
   \caption{Visualization by A-MIL on BACH dataset. } 
  \label{fig:bach_images}
\end{figure}   

We repeated the same experiments on BACH dataset for two-class classification and localization. We used non-overlapping patches of size $124 \times 124$ to form a bag from a single image. A bag contains $192$ instances for this experiment. A-MIL classifier achieved more than $80 \%$ accuracy on the test dataset. Figure \ref{fig:bach_images} shows localization results on a BACH dataset. Localization on BACH dataset also gives better attention to the epithelial region, which is important for tumor analysis.

\section{Conclusion}
We have shown the importance of attention mechanism, as it accurately highlights the region of interests and gives improved results in localization as compared to other architectures. Our approach not only provides the final diagnosis but also shows the meaningful interpretation of ROIs, which is difficult and extremely important in many clinical applications.

\section*{Acknowledgment}
Authors would like to thank Nvidia Corporation for donation of GPUs used for this research.

\bibliography{citations}
\bibliographystyle{IEEEtran}

\end{document}